\documentclass[letterpaper, 10 pt, conference]{ieeeconf}  

\IEEEoverridecommandlockouts                              

\usepackage{cite}
\usepackage{amsmath,amssymb,amsfonts}
\usepackage{algorithm}
\usepackage[noend]{algpseudocode}
\usepackage{graphicx}
\usepackage{textcomp}
\usepackage{array}
\usepackage{xcolor}
\usepackage{subcaption}
\usepackage{changepage}
\def\BibTeX{{\rm B\kern-.05em{\sc i\kern-.025em b}\kern-.08em
    T\kern-.1667em\lower.7ex\hbox{E}\kern-.125emX}}

\DeclareCaptionSubType[alph]{figure}
\makeatletter
\renewcommand\p@subfigure{\thefigure}
\renewcommand\thesubfigure{(\@alph\c@subfigure)}
\makeatother

\title{\LARGE \bf
Graph-based Simultaneous Coverage and Exploration Planning for Fast Multi-robot Search
}

\author{Indraneel Patil, Rachel Zheng, Charvi Gupta, Jaekyung Song, Narendar Sriram and Katia Sycara$^{*}$
\thanks{$^{*}$All authors are with the Robotics Institute at Carnegie Mellon University, Pittsburgh, PA. Emails:
        {\tt\footnotesize \{ipatil, rachelzh, cagupta, jaekyuns, narendas, sycara\}@andrew.cmu.edu}}%
}

\begin{document}

\maketitle
\thispagestyle{empty}
\pagestyle{empty}
\pagenumbering{arabic}

\begin{abstract}
In large unknown environments, search operations can be much more time-efficient with the use of multi-robot fleets by parallelizing efforts. This means robots must efficiently perform collaborative mapping (exploration) while simultaneously searching an area for victims (coverage). Previous simultaneous mapping and planning techniques treat these problems as separate and do not take advantage of the possibility for a unified approach. We propose a novel exploration-coverage planner which bridges the mapping and search domains by growing sets of random trees rooted upon a pose graph produced through mapping to generate points of interest, or tasks. Furthermore, it is important for the robots to first prioritize high information tasks to locate the greatest number of victims in minimum time by balancing coverage and exploration, which current methods do not address. Towards this goal, we also present a new multi-robot task allocator that formulates a notion of a hierarchical information heuristic for time-critical collaborative search. Our results show that our algorithm produces 20\% more coverage efficiency, defined as average covered area per second, compared to the existing state-of-the-art. Our algorithms and the rest of our multi-robot search stack is based in ROS and made open source\footnote{https://github.com/MRSD-Team-RoboSAR/robosar}.


\end{abstract}

\section{INTRODUCTION}

Multi-robot search in time-critical scenarios is a widely studied topic because of its many applications in pursuit-evasion, landmine detection, target tracking and search and rescue \cite{intro1,intro2}. We consider specifically the problem of using multi-robot fleets for autonomous search for victims in unknown environments for disaster relief. Robots must 1) create an online representation of the environment for exploration, 2) search for victims by covering the explored areas, and 3) generate traversable plans to complete the search in minimal time.

A popular method for addressing online multi-robot search in unknown environments involves dividing the problem into two distinct sub-problems: coverage and exploration \cite{splitcovexplore1,splitcovexplore2,splitcovexplore3,splitcovexplore4}. Exploration is defined as robots mapping unknown space, and coverage is defined as robots sweeping the mapped free space with a sensor to cover the whole area. This distinction is often helpful since exploration is very fast with laser based sensors like LIDARs which have greater range whereas coverage is more effective with vision based sensors like cameras. 

But this hard distinction between these two subproblems and tackling them independently with different algorithms and different environmental representations is very computationally costly and infeasible for online search.

Further, the dynamic nature of the map produced by SLAM (Simultaneous Localization and Mapping) produces several challenges for the above subproblems. The most popular exploration approach is frontier-based exploration in which robots move to the boundary between open space and uncharted terrain known as frontiers \cite{intro7} \cite{intro75}. Rapidly-exploring Random Trees (RRT) are used commonly to detect these frontiers because of their computational efficiency \cite{intro8} but these global trees suffer heavy pruning in a dynamic map and therefore need continuous regrowing. A local tree can be grown rooted on the robot's current position, but this results in the need to continuously regrow the tree and thus myopic behavior where frontiers far from the robots remain undetected.

\begin{figure}[t]
    \centering
    \includegraphics[width=9cm]{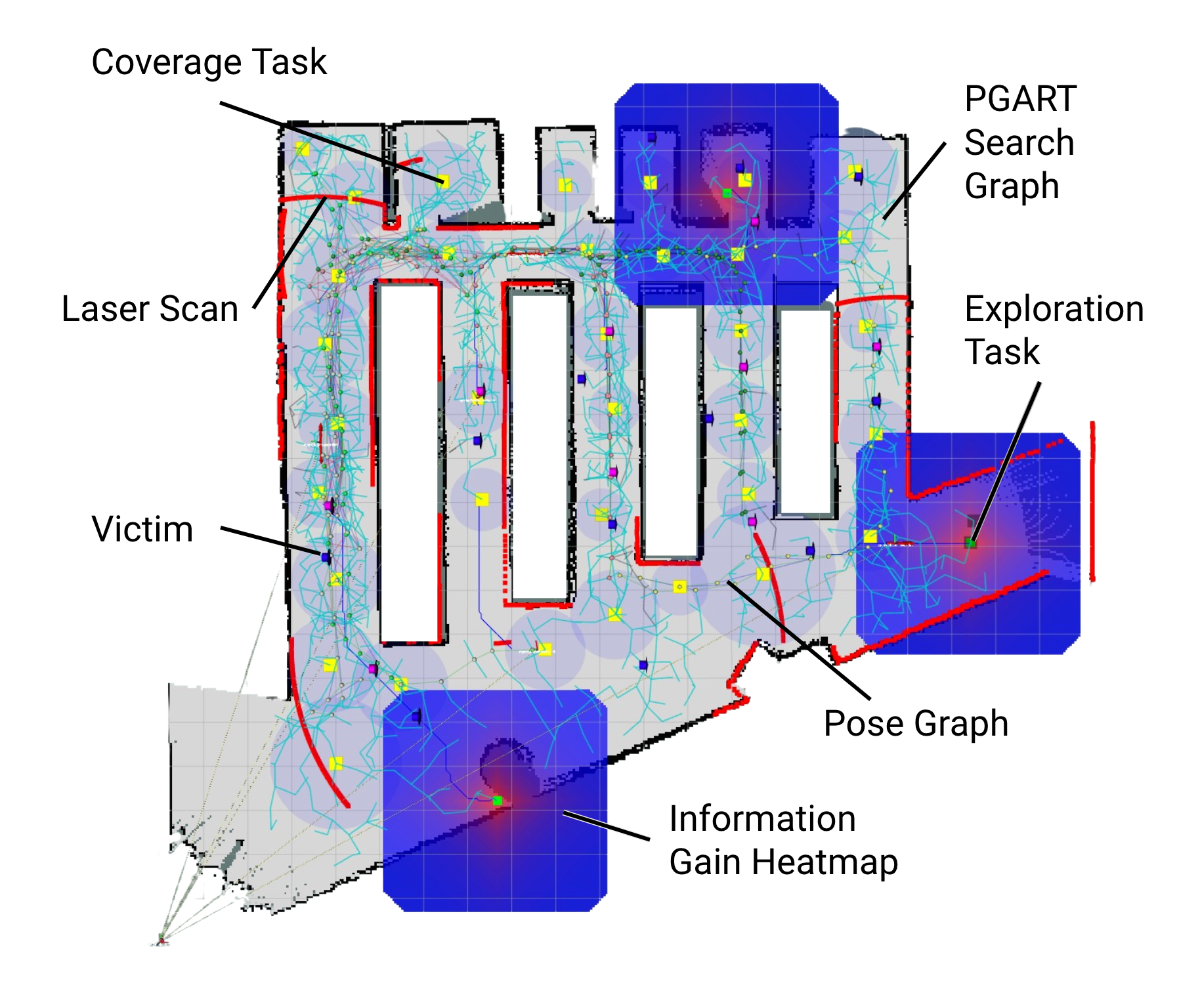}
    \caption{ The occupancy grid map along with the detected victims (dark blue points) in an indoor simulated environment during a search mission by a fleet of five robots. The green random trees are grown by the PGART planner on the pose graph produced by our multi-robot SLAM algorithm. The green and yellow nodes on the trees are the frontier and coverage nodes respectively. The red-blue information heatmap shows the relative rewards of each task used by the HIGH task allocator. }
    \label{fig:galaxy}
\end{figure}

On the other hand, fast coverage planning needs to ensure sufficient coverage with minimal sensor configuration and no backtracking (revisiting already covered areas). Optimal coverage solvers are only possible with known environments \cite{intro9} and hence not feasible in an online setting.  Most popular coverage methods based on cell decomposition\cite{intro10}, Voronoi diagrams \cite{int11} or wavefront algorithms are stateless and need very expensive data association to keep track of visited areas after loop closure. A proposed state-full solution to keep track of visited areas is using search graphs\cite{4}, but it still does not address retaining information after loop closures.

Motivated by these shortcomings and the need for a unified algorithm for both coverage and exploration planning, we propose our first contribution of a new online coverage-exploration planner called Pose Graph Aware Random Trees (PGART). PGART uses random trees grounded on an underlying pose graph generated by a graph-based SLAM algorithm \cite{intro11} to generate exploration and coverage tasks, shown in Figure \ref{fig:galaxy}. With this we avoid redundant graph construction unlike other graph-based search methods \cite{gbs1,gbs3,gbs6} and also make the information graph adapt to dynamic map updates.
 
Next, these coverage and exploration tasks must be allocated to the robots in the fleet to minimize the time taken to search the environment for victims. Next Best View (NBV) algorithms compute a sequence of viewpoints (while trading off information gain vs distance cost) until an entire scene or the surface of an object has been observed by a sensor\cite{IT3,IT5,IT6,IT9}. A previous solution, greedy Next Best View Planner (NBVP) \cite{IT4}, does not distinguish between coverage and exploration tasks and assign them with equal priority. This approach results in slow exploration and the lack of a “big picture view” of the environment, leading to suboptimal coverage planning and allocation.

A priori knowledge of the information distribution of victims in the environment can be used to speed up search (ergodic search \cite{intro13} and adaptive sampling  \cite{intro14})  but availability of this information in advance or the ability to sample it is unrealistic in real world disaster relief scenarios.

To address this, we introduce our second algorithm: a new information heuristic task allocator for simultaneous coverage and exploration called Hierarchical Information Gain Heuristic (HIGH) allocator. The HIGH allocator quantifies the information gain of coverage and exploration tasks, as visualized in Figure \ref{fig:galaxy}, by defining a hierarchical heuristic tree and assigns the robots to the tasks with a custom reward function. The hierarchical heuristic tree acts as a stand-in for the unknown distribution of the locations of victims. 

We evaluate our algorithms using a fleet of ground robots in simulated and real world experiments. We show that our algorithm is 27.7\% more search efficient than RRT-exploration \cite{intro8} and 20.0\% better than NBVP \cite{IT4}, which are previous state-of-the-art methods. Main contributions of this paper are summarized as follows:
\begin{itemize}
\item We propose an online unified coverage-exploration planner for objects-of-interest search in an unknown environment that is directly tied to the underlying SLAM pose graph.
\item We present an information heuristic task allocator for estimating information utility of coverage and exploration tasks and biasing search towards high information areas to minimize search time.
\item We make our entire multi-robot search hardware-agnostic framework open source to help further multi-robot research.
\end{itemize}

\begin{figure}[t]
    \centering
    \includegraphics[width=9cm]{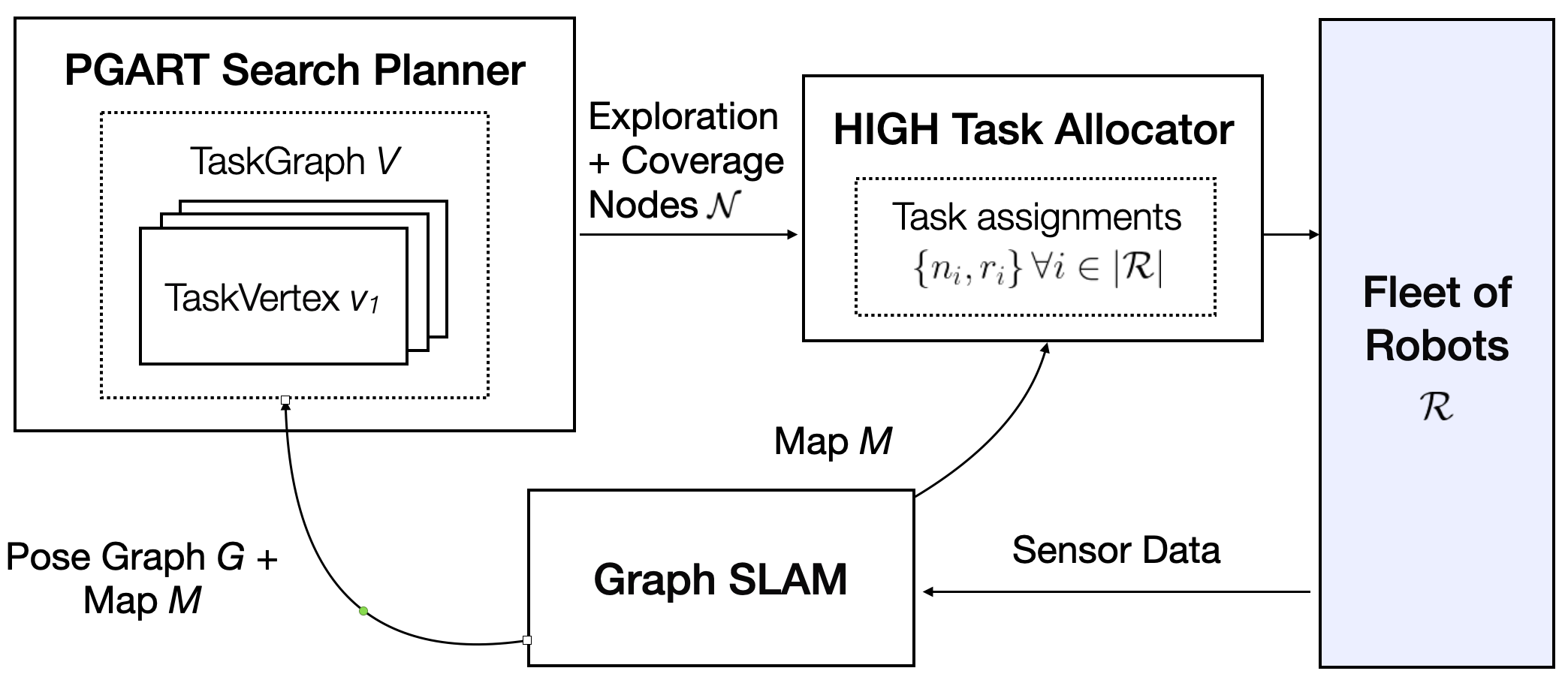}
    \caption{High-level system diagram which shows how the search planner and the task allocator are connected to the graph-based SLAM algorithm, to each other, and to the fleet of robots. }
    \label{fig:system}
\end{figure}
\section{Problem Description}
Consider a multi-robot search scenario in an unknown environment with a certain number of victims with unknown locations. Two key characteristics of the scenario are 1) the environment is large and 2) the time for rescue is limited. The problem can be thought of as a time-constrained optimization problem. Resources are the number of robots. The objective is to search for and locate the victims in the environment autonomously in minimum time.

The environment is defined as a 2D area $A \subset \mathbb{R}^2$ which denotes obstacles and free space. $A_{free} \subset \mathbb{R}^2$ comprises of all the 0 values in $A$ and $A_{obs} \subset \mathbb{R}^2$ comprises of all 1 values in $A$. The area $A$ is bounded by a rectangular geofence $F = \{(x_1,y_1), \hdots (x_4,y_4)\}$ which is the estimated boundary of the search area in the environment to produce $A_{bounded}$. The $n$ robots denoted by set $\mathcal{R}=\{r_1, r_2, r_3, .. r_n\}$ start together at an origin frame inside $A_{bounded}$ defined by $O = \{(x_0 , y_0)\}$.

The 2D occupancy grid map $M$ which represents $A_{bounded}$ is generated online during the operation and used for localization and navigation. Each point $x \in \mathbb{R}^n$ in $M$ has an occupancy $o_M(x) \in [0, 1]$. The desired localization of the victims in the environment can be defined by set $\mathcal{L} = \{\boldsymbol{l_0}, \boldsymbol{l_1} … \boldsymbol{l_{n_v-1}}\}$, Where each $\boldsymbol{l_i}= (x,y)$ is the 2D coordinate of the victim with respect to frame $O$ and $n_v$ is the number of victims found.

\section{Approach}
 Our work builds on top of key results from NBV algorithms, algorithms which use underlying graph data structures, and those which bridge SLAM and search \cite{gbs2}. Figure \ref{fig:system} depicts the high-level architecture of the system. The Graph SLAM module generates and updates a pose-graph and map, which serve as inputs to the PGART search planner and HIGH task allocator. The PGART search planner generates exploration and coverage nodes. The HIGH task allocator assigns these nodes to the available robots in order to maximize the overall reward. In this section, we provide a detailed description of the approach taken for each module. 

\subsection{Graph-based SLAM} \label{SLAM}
Our multi-robot SLAM implementation is built off of \texttt{slam\_toolbox} \cite{intro11}, which constructs a pose graph $\mathcal{G} = (X, E)$ whose nodes $X$ correspond to the robot's estimated pose and sensor measurement $Z$ at each time step, and edges $E$ are the spatial constraints between poses derived from odometry. An occupancy grid map M is also created where each cell $m \in M$ is either free, occupied, or unknown.

In this work, we extend \texttt{slam\_toolbox} to work with multiple robots by maintaining a central pose graph to which all robots' nodes and constraints are added. The pose graph is initialized with a global origin $O$ defined at the first robot $r_1$. Other robots are added to the pose graph using laser scan matching. When similar features are detected, for example via laser scan matching, loop closure occurs and the pose graph is optimized.

\subsection{PGART Search Planner} 

We consider the scenario where the sensor model of the search sensor (RGB camera) is different from the sensor model of the exploration sensor (laser scanner). Thus, to both map the environment and detect victims, our planner needs to simultaneously generate exploration and coverage nodes. The planner needs to handle a constantly changing map being built throughout exploration and still keep a track of searched areas to avoid redundant effort.

\subsubsection{\textbf{Tree Generation}} \label{TreeGenerate}

Consider the set of poses $X = (x_1,x_2,\ldots, x_n)$ from the pose graph $G$. We initialize a new data structure called a TaskGraph $V$ as visualized in Figure \ref{fig:system}. Task vertices are defined as $(v_1, v_2, \ldots, v_n) \in V$. Each task vertex extends a Rapidly-exploring Random Tree (RRT) \cite{intro12} rooted at $x_i$ to give the tree the “pose aware” functionality which we will discuss in section \ref{TreeUpdate}. Every tree $\mathcal{T}_i$ defines a set of randomly sampled 2D configurations or nodes $N_i = (n_i^1, \ldots, n_i^m) $ from the $A_{free}$ connected by collision-free edges $E_i$. We can define a union set of all nodes from every tree as $N_u = \bigcup_{i=1:n} N_{n_i}$. Assuming the RRT properties of probabilistic completeness, given sufficient time the set of all nodes $N_u$ gives us complete connectivity of the free space $A_{free}$ inside $A_{bounded}$ which the robots can use to find the NBVs in the environment.

The key motivation in the proposed tree expansion strategy, outlined in Algorithm \ref{pgart-alg}, is to maintain small trees $\mathcal{T}i$ around root pose $x_i$ to avoid tree pruning or regrowth, while still ensuring global connectivity of $A_{free}$ using a set of nodes $N_u$. We do this by growing the tree $\mathcal{T}_i$ from the closest task vertex $v_i$ to a randomly sampled configuration $n_{rand}$ using the standard RRT expansion strategy. If this fails, we find a node $n_{k_{closest}}$ from $N_u$ and then grow the tree $\mathcal{T}_k$ instead.

Our TaskGraph expansion strategy keeps the trees small and balanced which prevents pruning during map updates but also provides thorough expansion into $A_{free}$. 

\subsubsection{\textbf{Dynamic Tree Pose Update}} \label{TreeUpdate}
Since the pose graph $G$ and poses $x_i$ are subject to real-time optimization based on new sensor data from the robots as discussed in section~\ref{SLAM}, we dynamically update every task vertex $v_i$ to keep the TaskGraph inside $A_{free}$ to prevent excessive pruning. 

\begin{algorithm}[t]
    \caption{TaskGraph Expansion}\label{pgart-alg}
    \textbf{Input:} Unoccupied configurations $A_{free}$, TaskGraph $V$, Nodes $N_u$, extension distance $\eta$. \\
    \textbf{Output:} TaskGraph $V$
    \begin{algorithmic}[1]
        \State $n_{rand} \leftarrow$ \text{Sample}$(A_{free})$
        \State $v_{i} \leftarrow $ \text{ClosestTaskVertex}$(V, n_{rand})$
        \State $n_{i_{closest}} \leftarrow$ \text{ClosestNode}$(v_{i}, n_{rand})$
        \State $n_{new} = $ \text{Extend}$(n_{i_{closest}}, \mathcal{T}i, n_{rand}, \eta)$
        \If{\text{CollisionFree}$(n_{new}, n_{i_{closest}}, M)$}
        \State \text{UpdateRRT}$(\mathcal{T}i, n_{new})$
        \Else
        \State $n_{i_{closest}} \leftarrow$ \text{ClosestNode}$(N_u, n_{rand})$
        \State $n_{new} =$ \text{Extend}$(n_{i_{closest}}, \mathcal{T}i, n_{rand}, \eta)$
        \If{\text{CollisionFree} $(n_{new}, n_{i_{closest}}, M)$}
        \State \text{UpdateRRT}$(\mathcal{T}i, n_{new})$
        \EndIf
        \EndIf
        \State \Return TaskGraph $V$
    \end{algorithmic}
\end{algorithm}

\begin{algorithm}[t]
\caption{Exploration Node Filtering}\label{exploration-filtering}
\textbf{Input:} Candidate exploration nodes $N_{candidate}$, previous frontier nodes $N_{frontier}$, laser range $\beta$, information gain threshold $\xi_{exploration}$, mean shift bandwidth $\epsilon$, map $M$.
\\
\textbf{Output:} Final set of frontier nodes $N_{frontier}$.
\begin{algorithmic}[1]
\State $N_{candidate} \leftarrow N_{candidate} \cup N_{frontier}$
\For{$n_i \in N_{candidate}$}
\State $I_f \leftarrow$ \text{InformationGain}$(n_i, \beta, M)$
\State $d \leftarrow$ \text{EuclideanDistance}$(n_i, G)$
\If{$I_f < \xi_{exploration} $}
\State $N_{candidate} \leftarrow N_{candidate} - n_i$
\EndIf
\EndFor
\State $centroids, clusters \leftarrow$ \text{MeanCluster}$(N_{candidate}, \epsilon)$
\State $N_{frontier} \leftarrow centroids$
\For{$c_i, n_i \in centroids, clusters$}
\If{not \text{LineOfSight}$(n_i, c_i, M)$}
\State $N_{frontier} \leftarrow N_{frontier} \cup n_i$
\EndIf
\EndFor
\State \textbf{return} $N_{frontier}$
\end{algorithmic}
\end{algorithm}

Each TaskVertex $v_i$ stores the parent pose $x_i$ as a map (origin frame $O$) to root transformation and each tree $\mathcal{T}_i$ is stored in the local frame of the pose $x_i$. When motion is detected in pose $x_i$ we execute a two-step \textbf{UpdateTaskVertex} function on tree $\mathcal{T}_i$ as follows: 
\begin{enumerate}
    \item Update the map to root transform with the $x_{i_{new}}$ in the task vertex ${v_i}$
    \item Start at the root node, recursively traverse the tree $\mathcal{T}_i$ and apply this new map to root transformation on each node to find the new map to node transform for each node in the tree.
\end{enumerate}

\begin{figure} [t]
\centering
  \includegraphics[width=3.5cm]{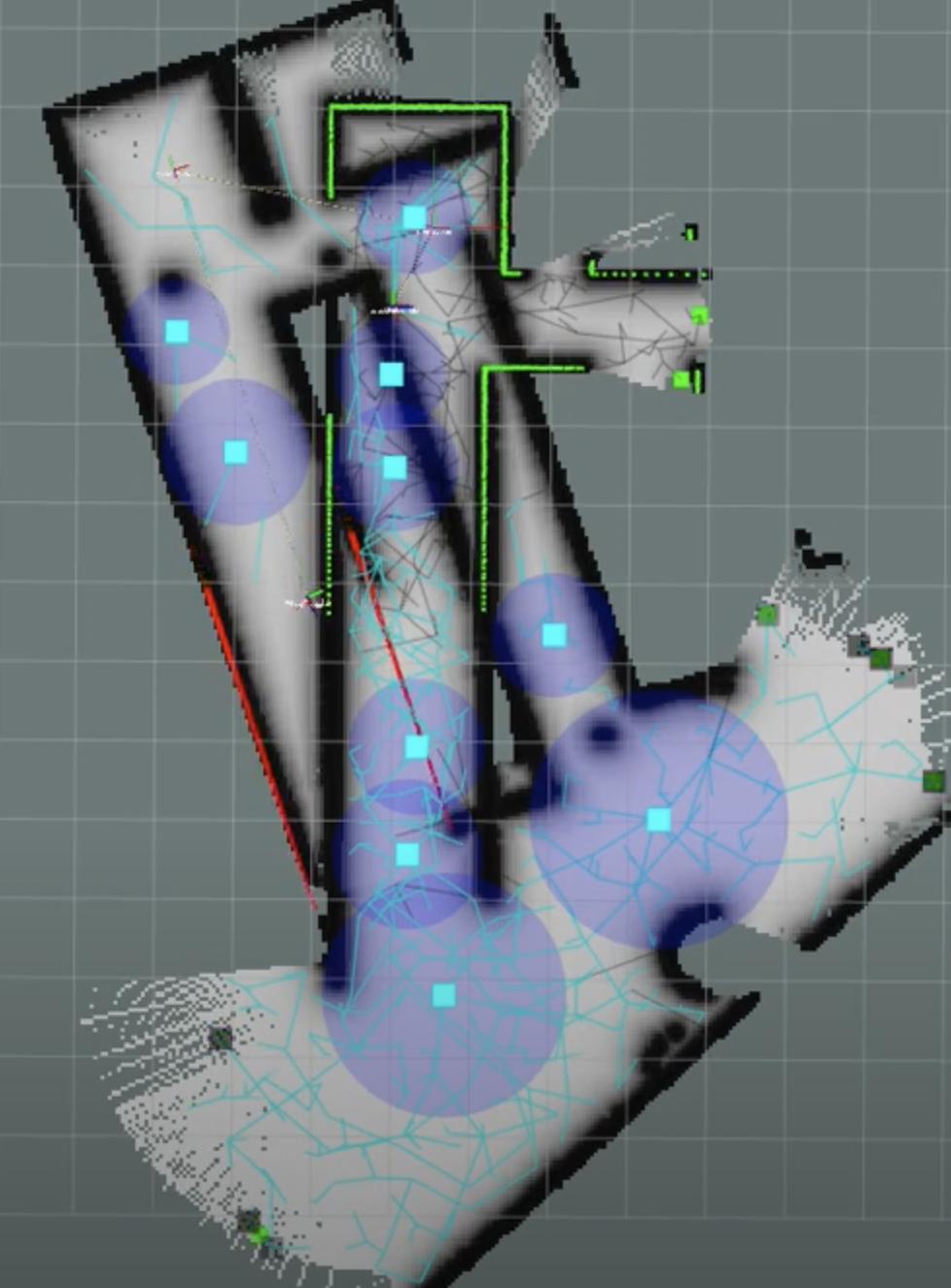}
  \includegraphics[width=3.5cm]{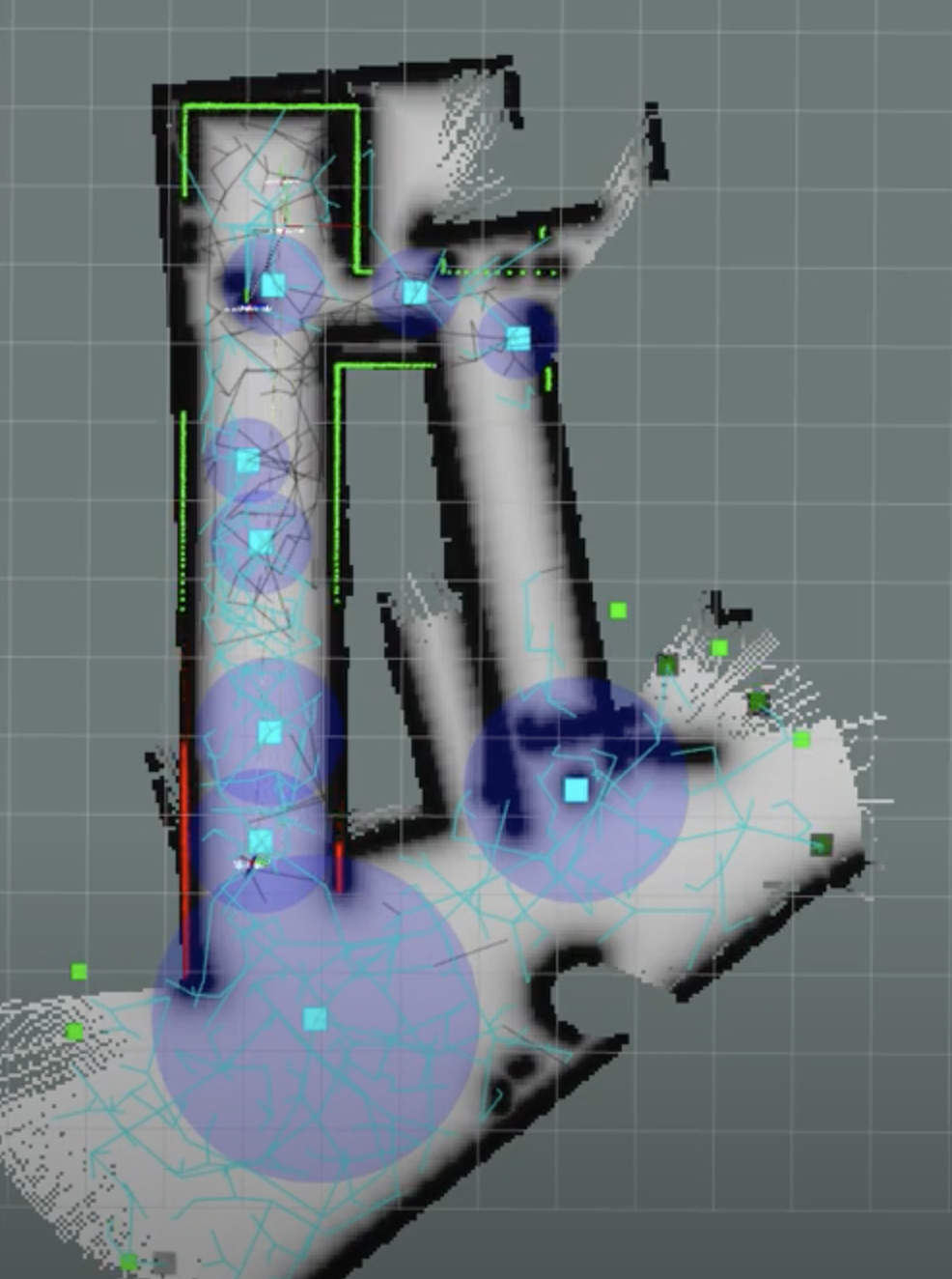}
\caption{(Left) PGART (cyan lines) before map update. Because of bad localization, there are two mapped versions of the same corridor on the left. Redundant coverage tasks (cyan squares) are generated. (Right) PGART after map update. The redundant tasks are now deleted, and branches intersecting obstacles (grey lines) are recursively pruned.}
\label{fig:PGART_before_after}
\end{figure}

This completes our dynamic tree pose transformation based on updates from the pose graph. Figure \ref{fig:PGART_before_after} shows an example of an update during a large loop closure, depicting the ability of our PGART planner to adapt with a dynamic map and remove redundant tasks.

 \subsubsection{\textbf{Generating Exploration Nodes}} 

We use the Task Graph $V$ to generate exploration nodes, defined as nodes from $N_u$ of the random trees $\mathcal{T}$ which connect from unknown to known space. 

Our method of exploration node generation, also known as frontier detection, using RRT closely resembles the work of \cite{intro8}. The tree generation procedure described above in section \ref{TreeGenerate} creates a set of candidate exploration nodes $n_{candidate}$. Node filtering is applied on the set of nodes to filter out low information, redundant, and other unreachable nodes as shown in Algorithm \ref{exploration-filtering} which are key improvements from \cite{intro8}.

In the first stage of filtering, nodes are filtered based on an information gain threshold $\xi_{exploration}$ and clustered together. The information gain of a frontier candidate is defined as the ratio between the area of continuous unknown cells \cite{intro75} within the sensor coverage area and the total sensor coverage area, as seen in equation \ref{info-gain-frontier}. This helps filter out old frontier nodes and also new nodes with low information gain. In the second stage, the candidates are clustered using mean-shift clustering to filter out the redundant candidates which are part of the same frontier front. We verify each frontier candidate in the cluster is within line-of-sight of its centroid. If not, it is removed from the cluster and saved as a separate frontier node. The threshold $\xi_{exploration}$ is a user set parameter.

The filtering process generates a final set of exploration nodes $N_{frontier}$ that have high information gain for effective exploration.

\begin{algorithm}[t]
\caption{Survival of the Fittest Coverage (SOTFC) Filter}\label{sotfc-alg}
\textbf{Input:} Set of trees ${\mathcal{T}}$, Sensor range $\xi_{range}$, Minimum visibility radius $\mathcal{V}_{r_{min}}$
\\
\textbf{Output:} Set of the best coverage nodes $N_{coverage}$
\begin{algorithmic}[1]
\For{$\mathcal{T}_i$ in ${\mathcal{T}}$}  \Comment{ Intra-tree filter}
\State $N_{candidate_i} \leftarrow$ Best coverage nodes of $\mathcal{T}i$ using Eq.~\ref{visibility-radius-coverage} with $R_A$ rays
\State $N_{coverage} \leftarrow N_{coverage} \cup N_{candidate_i}$
\EndFor
\For{$i$ in $|{\mathcal{T}}|$}  \Comment{ Inter-tree filter}
\For{$n_c$ in $N_{candidate_i}$}  
\If{$ \mathcal{V}_r(n_c) < \mathcal{V}_{r_{min}}$}
\State Remove $n_c$ from $N_{coverage}$
\EndIf
\For{$n_c^{'}$ in $N_{coverage} \setminus n_c$}
\If{$d(n_c,n_c^{'}) \leq 2r_c$}
\If{$\mathcal{V}_r(n_c) < \mathcal{V}_r(n_c^{'})$}
\State Remove $n_c$ from $N_{coverage}$
\Else
\State Remove $n_c^{'}$ from $N_{coverage}$
\EndIf
\EndIf
\EndFor
\EndFor
\EndFor
\State $N_{coverage} \leftarrow$ First $k$ nodes with highest $\mathcal{V}_r(N_{coverage})$
\end{algorithmic}
\end{algorithm}

\subsubsection{\textbf{Generating Coverage Nodes}}

From the Next-Best View planning literature, we define a good coverage node as a point inside $A_{free}$ which provides the most gain in surface coverage and uncertainty reduction \cite{IT5}. For this, we define a simplistic circular sensor model with range  $\xi_{range}$, and $R_A$ as the set of rays cast radially outwards from the node. The visibility radius of surface coverage of a particular node is the total uninterrupted free space observable from that node, and is defined as:

\begin{equation} \label{visibility-radius-coverage}
    \mathcal{V}_r  = \pi r_c^2, \quad \text{where} \quad r_c = min \left( max |R_A| ,\xi_{range} \right)
\end{equation}

We designed a computationally efficient SOTFC (survival of the fittest coverage) filter as described in Algorithm \ref{sotfc-alg} which uses this formulation of visibility radius. SOTFC is a two-level filter: on the first level, the intra-tree filter produces the best coverage nodes from each random tree $\mathcal{T}_{i}$. On the second level, a final inter-tree filter produces a final set of coverage nodes $N_{coverage}$. This makes SOTFC efficient with good real-time performance since it avoids unnecessary intra-tree computation. For example both online dynamic pose update from section \ref{TreeUpdate} (since that moves trees as a whole) and the addition of a new TaskVertex $v_i$ with tree $\mathcal{T}_i$ only need inter-tree filtering.

A caveat in coverage node generation is that due to the online nature of tree generation discussed in \ref{TreeGenerate} the quality of coverage nodes is sub-optimal at the start but improves over time as enough configurations are sampled. But due to the probabilistic completeness property of RRT, the set $N_{coverage}$ of coverage nodes always converges to near-optimal coverage configurations given enough time. Since this is not possible in time-critical search scenarios, we propose the SOTFC filter as an anytime coverage algorithm for dynamic unknown environments.

\subsection{HIGH Task Allocator} \label{TaskAllocator}

The task allocator takes the set of exploration and coverage nodes as input task candidates, defined as $\mathcal{N} = N_{frontier} \cup N_{coverage}$. It assigns each robot to these tasks based on a custom reward function in each assignment round. A round of assignments occurs whenever any robot completes a task, or the assignment period is reached. Compared to previous methods like NVBP, HIGH additionally utilizes an information heuristic approach that weighs exploration vs coverage importance. It then exploits these objectives to balance exploration vs coverage. HIGH maximizes a reward function composed of the following four objectives:

\subsubsection{\textbf{Action Cost}} We define the action cost as $D(n, x_r)$, the A* path distance to each task $n_i \in \mathcal{N}$ from robot position $x_r$. We want to minimize this cost.

\subsubsection{\textbf{Utility}} If the task is within line-of-sight range to a currently assigned task in $\mathcal{N}_a$, then its utility is decreased. This discourages different robots from being assigned tasks that are very close to each other. 

\begin{equation}
  U(n)=
  \begin{cases}
    1, \quad \text{if} \quad \prod_{n_i \in \mathcal{N}_a} \textbf{raycast}(M, n, n_i) == 1 \\
    \frac{min(1, \textbf{euclidean}(n, n_{a}^{closest}))}{\beta}, \quad \text{otherwise}
  \end{cases}
\end{equation}

where \textbf{raycast} returns 1 if there is no collision and 0 otherwise, and $\beta$ is the laser scanner range.

\subsubsection{\textbf{Information Gain}} 

\begin{algorithm}[t]    
    \caption{HIGH Task Allocator}\label{Alg-Decap}
    \textbf{Input:} Map $M$, Robots $R$, Unassigned Tasks $\mathcal{N}$, Camera range $\gamma$, Laser range $\beta$ \\
    \textbf{Output:} Task assignments $\Phi$ for each robot
    \begin{algorithmic}[1]
        \State $\mathcal{N}_p = \emptyset$
        \State $\Phi = \emptyset$
        \State $a \leftarrow $ MappedArea$(M)$
        \For{Robot $r \in R$}  
        \If{$\mathcal{N} != \emptyset$}
        \State $\mathcal{N}' \leftarrow $ SampleTasks$(T)$
        \State $n_r = \underset{n \in \mathcal{N}'}{\max} \hspace{4pt} I(n, \gamma, \beta) W(a)(\lambda U(n, \beta) – D(n, x_r))$
        \vspace{4pt}
        \State $\mathcal{N}_p \leftarrow \mathcal{N}_p \cup n_r$
        \State $\mathcal{N} \leftarrow \mathcal{N} - n_r$
        \EndIf
        \EndFor
        \State $\{i, r_i\} \leftarrow$ JonkerVolgenantAssignment$(\mathcal{N}_p, R)$
        \State $\Phi \leftarrow \Phi \cup \{n_i, r_i\} \hspace{5pt} \forall i$
        \State \Return $\Phi$
    \end{algorithmic}
\end{algorithm}

The information gain is defined as $I(n)$, the percentage of unknown cells that are within the visibility region of task candidate $n_i \in \mathcal{N}$. Tasks with high information gain should be prioritized in assignments.

For exploration tasks, information gain $I_f(n)$ is defined as the percent of continuous unknown cells within the laser scanner range $\beta$, which is calculated using a flood-fill algorithm. For coverage tasks, information gain $I_c(n)$ is defined as the percentage of free cells within camera sensor range $\gamma$ that have not yet been covered. However, it is non-trivial to keep a track of which free cells have been covered in a dynamic occupancy map. Thus we use a heuristic for the coverage task information gain: the Euclidean distance to the nearest pose graph node. If a coverage task is farther from the pose graph, then it is less likely to have been previously covered by the camera sensor, and should therefore have a higher information gain. We then normalize this value with the camera sensor range $\gamma$. Therefore, if a coverage task is more than $\gamma$ distance away from the pose graph, then it will have an information gain of 1. Else, the information gain heuristic is scaled linearly. 

\begin{equation} \label{info-gain-frontier}
    I_f(n) = \frac{\textbf{flood\_fill}(n, M, \beta)}{\textbf{area}(\beta)}
\end{equation}

\begin{equation}
    I_c(n) = min \left( \frac{\textbf{euclidean}(n, v_{closest})}{\gamma}, 1 \right)
\end{equation}

where $v_{closest}$ is the closest pose graph vertex.

\subsubsection{\textbf{Exploration-Coverage Importance}} When we first start the mission, all discoverable information is encoded in the frontiers. So at first exploration tasks have higher information, but as mapping progresses coverage tasks should gain relative importance. We use a linearly decaying importance weight for exploration as a function of the percent of area explored $a$. The total area is approximated by the user-specified geofence $F$. As more area is explored, the exploration weight $W_f(a)$ decreases while the coverage weight $W_c(a)$ increases. 
This means that the geofence area definition is significant in the prioritization of exploration vs coverage in early exploration. If the geofence area is much larger than the actual enclosed environment, then the exploration weight will decay very slowly, and the robots will highly prioritize visiting exploration tasks until there are only coverage tasks left.

\begin{equation}
    W_c(a) = min\left(\frac{a}{\textbf{area}(F)}, \hspace{2pt} 1 \right)
\end{equation}
\begin{equation}
    W_f(a) = 1 - W_c(a)
\end{equation}

The final HIGH reward function we want to maximize for a single robot is calculated hierarchically as:

\begin{equation}\label{HIGH_reward_fn}
   \max_{n \in \mathcal{N}} \hspace{5pt} I(n) W(a)(\lambda U(n) – D(n, x_r))
\end{equation}

\begin{table*}[!t]
\begin{center}
\caption{\label{tepper-sim-table} Experiment Results}
\begin{tabular}{ | c | c | c | c | c | c | c | c | c | c | c | c | c |} 
    \hline
    \multicolumn{1}{|c}{} & \multicolumn{3}{|c|}{Tepper Simulation} & \multicolumn{3}{c|}{Tepper Real} & \multicolumn{3}{c|}{World 1} & \multicolumn{3}{c|}{World 2}\\
    \hline
    \hline
    & $\epsilon$ & SST & \% Victims & $\epsilon$ & SST & \% Victims & $\epsilon$ & SST & \% Victims & $\epsilon$ & SST & \% Victims\\
  \hline
  HIGH & \textbf{0.45} & \textbf{2637} & \textbf{96.7}\% & \textbf{0.40} & \textbf{3472} & \textbf{91.7\%} & \textbf{0.25} & \textbf{2343} & \textbf{87.5\%} & \textbf{0.49} & \textbf{3689} & \textbf{87.5}\%\\ 
  \hline
  NBVP & 0.36 & 4009 & 91.7\% & 0.32 & 4969 & 83.3\% & 0.22 & 4469 & 52.5\% & 0.36 & 6213 & 63.3\%\\ 
  \hline
  RRT Exploration & 0.32 & 5182 & 71.1\% & - & - & - & \textbf{0.25} & 3538 & 70.0\% & 0.46 & 5765 & 68.3\%\\ 
  \hline
\end{tabular}
\end{center}
\end{table*}

\begin{figure*}[h]
    \begin{adjustwidth}{-1cm}{-1cm}
    \centering
    \begin{subfigure}[b]{6cm}
        \centering
        \includegraphics[width=\linewidth]{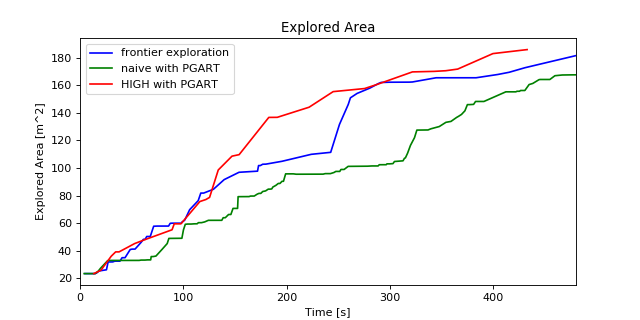}
        \caption{}
        \label{fig:FCa}
    \end{subfigure}
    \begin{subfigure}[b]{6cm}
        \centering
        \includegraphics[width=\linewidth]{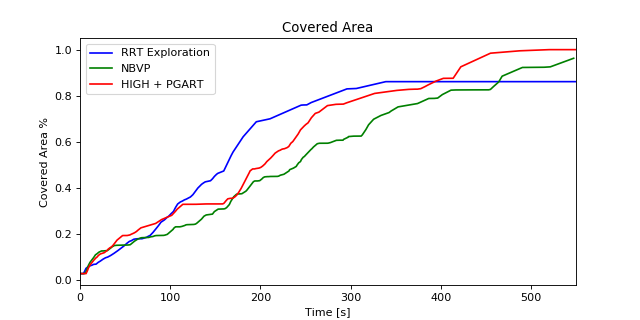}
        \caption{}
        \label{fig:coverage}
    \end{subfigure}
    \begin{subfigure}[b]{6cm}
        \centering
        \includegraphics[width=\linewidth]{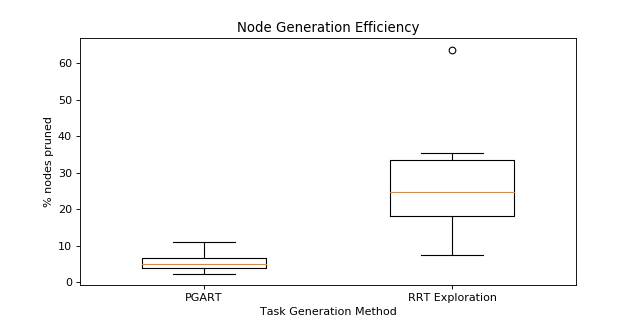}
        \caption{}
        \label{fig:FCb}
    \end{subfigure}
    \begin{subfigure}[b]{6cm}
        \centering
        \includegraphics[width=\linewidth]{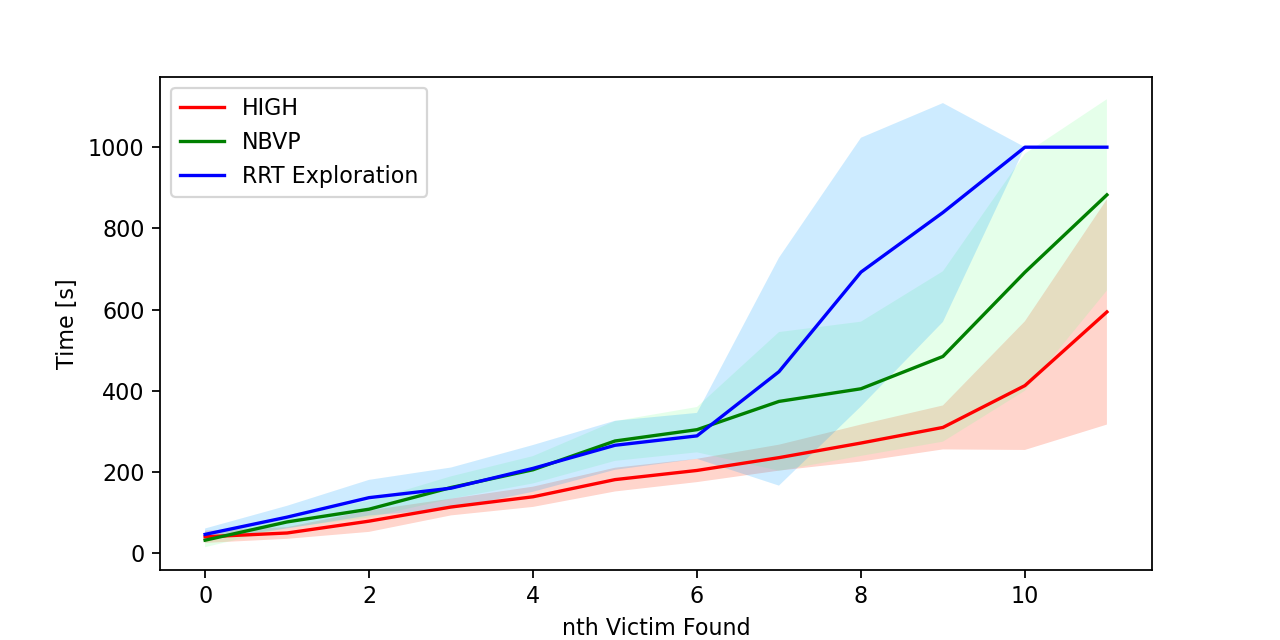}
        \caption{}
        \label{fig:metrics-SSTa}
    \end{subfigure}
    \begin{subfigure}[b]{6cm}
        \centering
        \includegraphics[width=\linewidth]{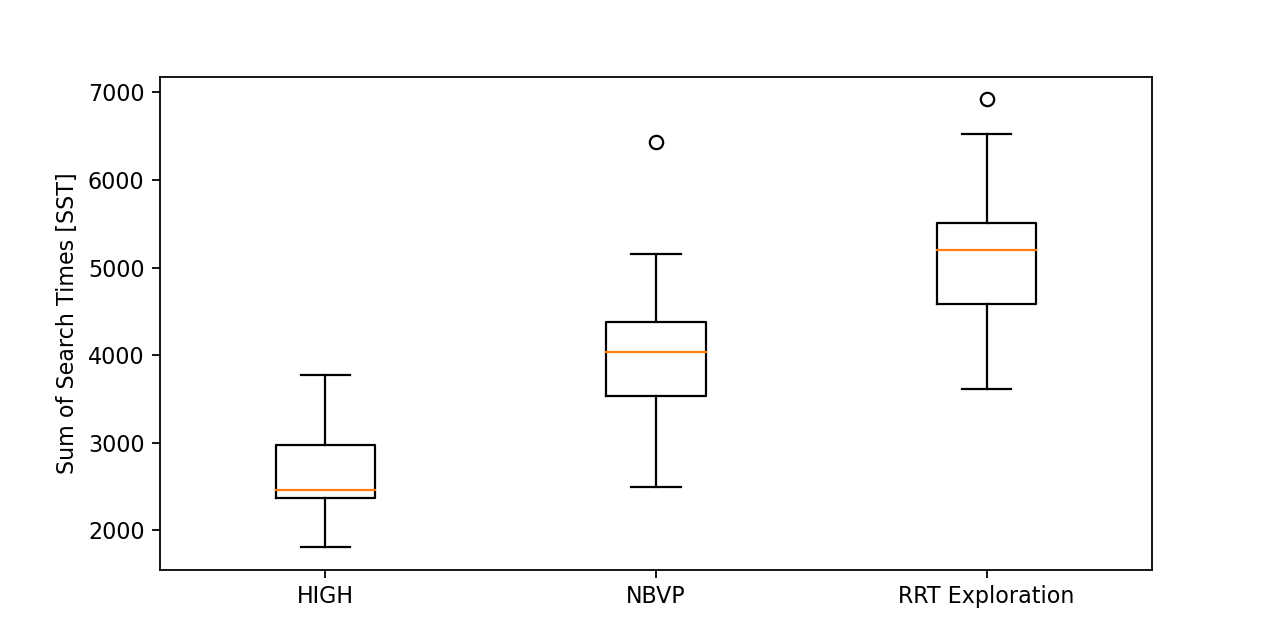}
        \caption{}
        \label{fig:metrics-SSTb}
    \end{subfigure}
    \begin{subfigure}[b]{6cm}
        \centering
        \includegraphics[width=\linewidth]{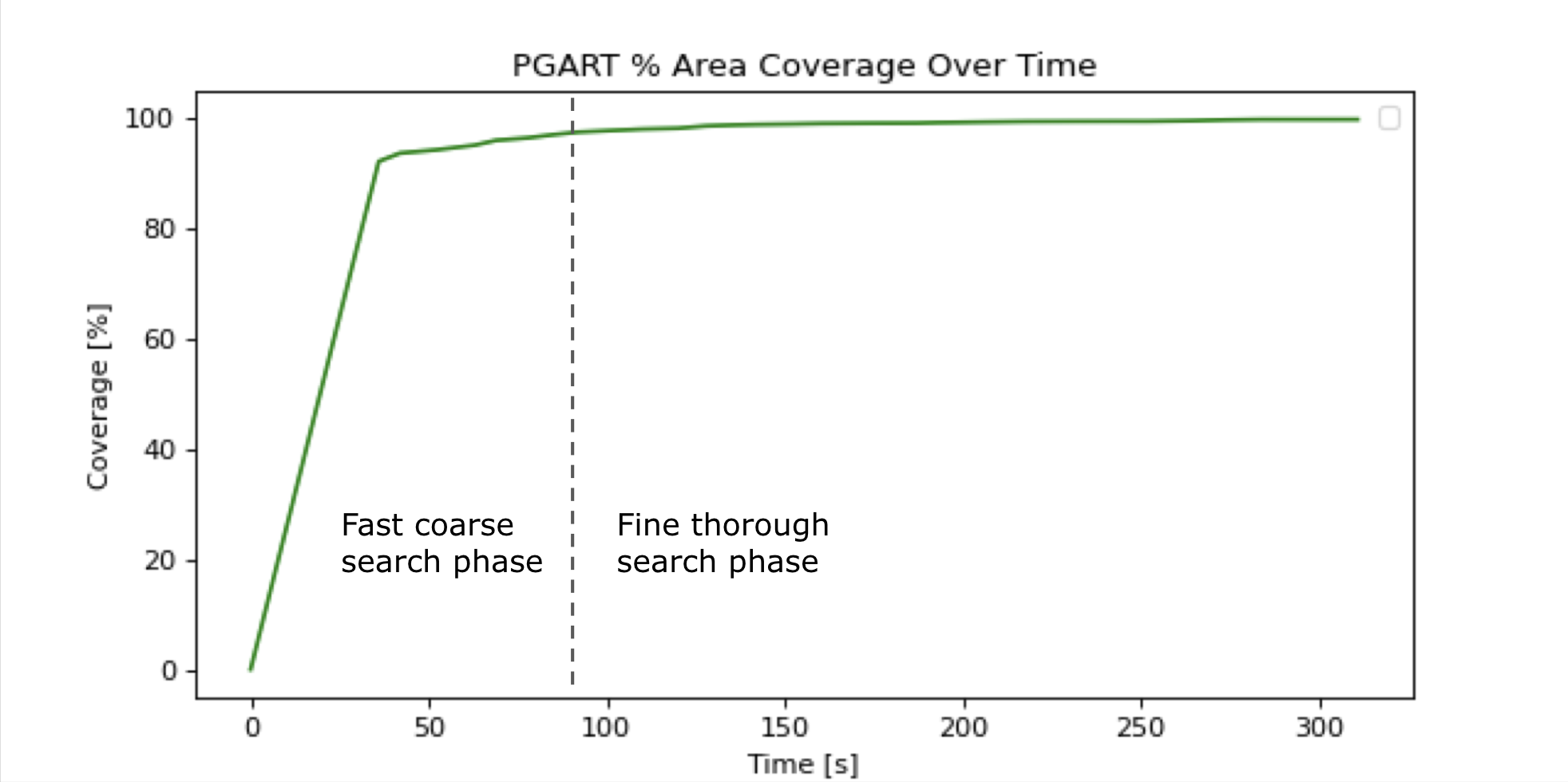}
        \caption{}
        \label{fig:completeness}
    \end{subfigure}
    \end{adjustwidth}
    \caption{Results averaged from 20 simulation runs in the Tepper Building world. (a) shows the rate of exploration from a representative run. (b) shows the rate of coverage from a representative run. (c) shows the percent of nodes pruned for PGART and RRT Exploration averaged over 20 simulation runs. (d) shows the average time in seconds the nth victim was found. (e) shows the average SST for each task allocation method, and shaded regions are the standard deviation. (f) depicts that the \%coverage area of PGART planner tends to 100\% with increasing time. }
    \label{fig:metrics}
\end{figure*}

Because it is very computationally expensive to compute A* on a large map for all tasks, we define a \textbf{SampleTasks} function that uses weighted random choice with information gain as the weights to sample a set of tasks to consider during an assignment round. We get the highest priority task for each robot in a round-robin fashion by selecting the unassigned task with the highest reward according to Equation \ref{HIGH_reward_fn}. After calculating the highest priority tasks to assign next, we then optimally assign the tasks to each robot based on distance cost using the Jonker-Volgenant algorithm \cite{taskalloc}. The total distance, and therefore time, needed to be traveled for all robots is then minimized.


\section{Experiments and Results} 

\subsection{System Summary and Setup}

 We evaluate the performance of our system in a real-world scenario using five Khepera IV robots and in three simulated indoor environments using the ROS-Stage simulator, to demonstrate the generalizability and robustness of our approach. The test setups comprise of 12 AprilTags as victims in the environment distributed randomly. Rectangular geofence $F$ is input by the user in order to define a bounded search area.

Our implementation of the PGART planner, the HIGH allocator, and the rest of the multi-robot stack capable of end-to-end autonomous search in any unknown environment is based on ROS. Our stack has been designed with a modular architecture, and can be used with real or simulated robots. This modular design makes our stack highly customizable and easy to use, making it one of few full stack solutions available for multi-robot systems.

\subsection{Evaluation Metrics}

We use 3 evaluation metrics:
\subsubsection{Coverage Efficiency} This is the average area viewed by the coverage sensor over a mission run (m$^2$/s), denoted as $\epsilon$. We end the mission once all victims are found, which is on average when 90\% coverage is reached. We set the mission time limit to two times our longest run.

\subsubsection{Percent of Victims Found} This is the percent of victims detected at the end of the mission.

\subsubsection{SST} The Sum of Search Times (SST) is the sum of the time taken to locate each victim $t_i$ and a penalty $P$ for undiscovered victims. The total penalty is calculated as $P = {(n_v  - n_f)*p_{max}}$ where $p_{max}$ is the maximum penalty for each undiscovered victim, $n_v$ is total number of victims, and $n_f$ is the number of found victims. For our experiments, $p_{max}$ is set as 1000 seconds. SST is then defined as:

\begin{equation}
    SST = \sum_{i=1:n_f} t_i + P
\end{equation}


We compare our proposed PGART + HIGH method to two previous state-of-the-art exploration algorithms: RRT Exploration \cite{intro8}, and NBVP \cite{IT4}. To ensure the reliability of the data, each method is repeated 20 times for each environment.

\subsection{Simulations}

We ran experiments in an indoor simulated environment, Tepper World, of size $17m \times 20m$. In the simulations, mapping was performed using laser range finders, whose sensor range is set to $\beta = 4m$ with FOV of $145$ degrees, the estimated hardware ranges. The maximum velocity of the robots is set to 0.3m/s, to mimic hardware conditions. Figure \ref{fig:galaxy} shows the resulting map and search graph from a representative run.

Over 20 simulation runs, our system on average detected 96.9\% of victims within the mission search time and achieved an average coverage efficiency of 0.45 as seen in Table \ref{tepper-sim-table}. Our method is able to cover the environment 20.0\% more efficiently than NBVP and 27.7\% than RRT Exploration. As seen in Figure \ref{fig:metrics}\subref{fig:coverage}, NBVP is able to eventually reach 100\% coverage but at a slower rate than our method, while RRT Exploration is able to cover a lot of area early in the mission but saturates once there are no more frontiers. On the other hand the PGART planner follows up the fast coarse search with a fine thorough search due to its anytime nature and is guaranteed to asymptotically reach 100\% coverage as seen in Figure \ref{fig:metrics}\subref{fig:completeness}.

Figure \ref{fig:metrics}\subref{fig:metrics-SSTa} depicts the time each nth victim was found, averaged over all runs. HIGH clearly finds each victim faster than NBVP or RRT exploration. The two baseline approaches have similar performance early in the mission since they both greedily visit the next closest task. However, the addition of coverage tasks greatly increases the efficiency of finding more victims later in the mission for NBVP. This is likely because those later victims are hidden in areas that were covered by the exploration sensor, but uncovered by the search sensor. RRT exploration was only able to find 71.1\% of victims which is significantly lower than both NBVP (91.6\%) and HIGH (96.9\%). Therefore, PGART is able to bridge this gap by generating more tasks to better cover the environment and increase the number of victims found. Additionally PGART only had a 5.3\% node pruning rate while RRT exploration had an average of 26.7\%, seen in Figure \ref{fig:metrics}\subref{fig:FCb}, allowing for more consistent tree expansion and task retention. 

Although HIGH only found 5.3\% more victims than NBVP, HIGH on average performed 34.2\% better than NBVP on the Sum of Search Times (SST) metric, seen in Figure \ref{fig:metrics}\subref{fig:metrics-SSTb}. This indicated that although both algorithms are able to find most victims within the mission runtime, HIGH is able to find them much faster, resulting in a significantly lower SST. Additionally, HIGH on average has a 47.1\% lower standard deviation of SST, meaning that its performance is much more consistent and repeatable than NBVP.

Figure \ref{fig:metrics}\subref{fig:FCa} depicts the rate of exploration. HIGH consistently is able to explore more area than NBVP throughout the entire mission. This shows HIGH is able to explore the same amount of area as RRT exploration, which only greedily maximizes exploration, while also being able to find more victims. Therefore, HIGH is able to balance both exploration and search, resulting in both a greater amount of victims found within mission runtime and lower victim detection times.

\begin{figure}[t]
\centering
    \includegraphics[width=4cm]{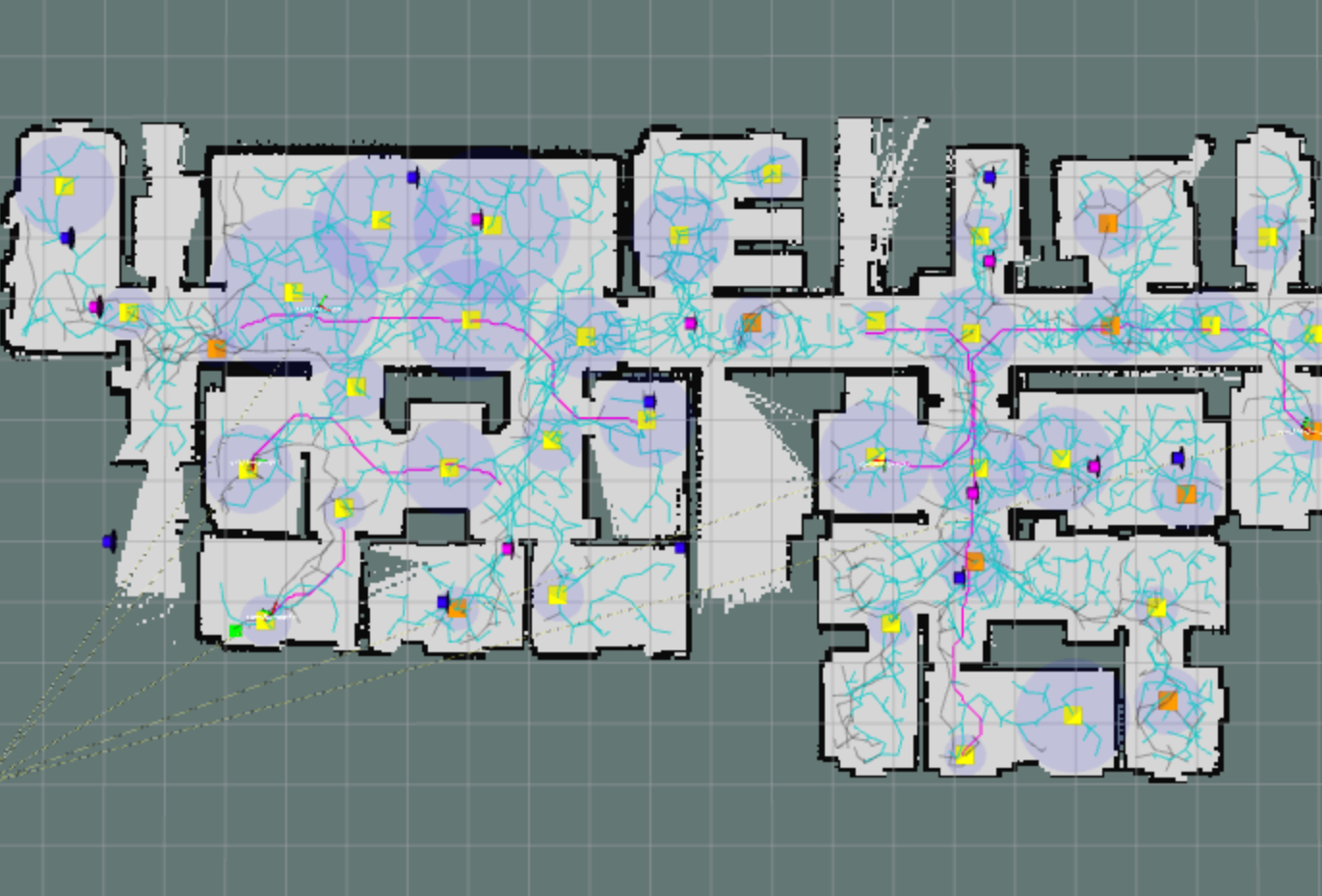}
    \includegraphics[width=3cm]{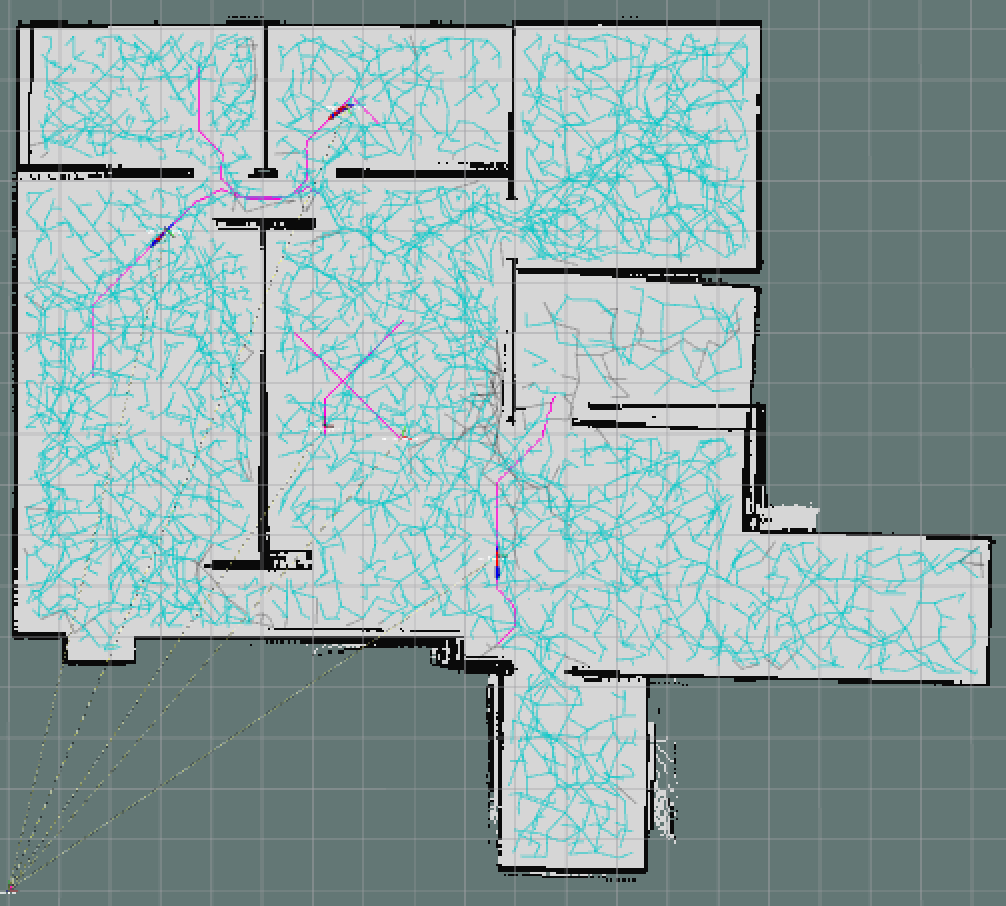}
    \caption{Additional simulation environments used in experiments. (left) World 1. (right) World 2.}
    \label{fig:test_worlds}
\end{figure}

To validate that our approach transfers to environments of varied structure, we also run experiments on two other worlds, depicted in Figure \ref{fig:test_worlds}. The results in Table \ref{tepper-sim-table} show that HIGH has the lowest SST and highest victim detection rate for both the worlds. However, the values of $\epsilon$ for World 1 are comparable for all the 3 algorithms. This is because it has narrow corridors and small rooms, which results in the area seen by the exploration and coverage sensors to be similar. For World 2, there is a significant improvement in $\epsilon$ for HIGH as compared to NBVP. This world has larger rooms and therefore experiences more coverage efficiency benefit from our method.

\subsection{Real experiments}

\begin{figure}[!t]
\centering
    \includegraphics[width=5cm]{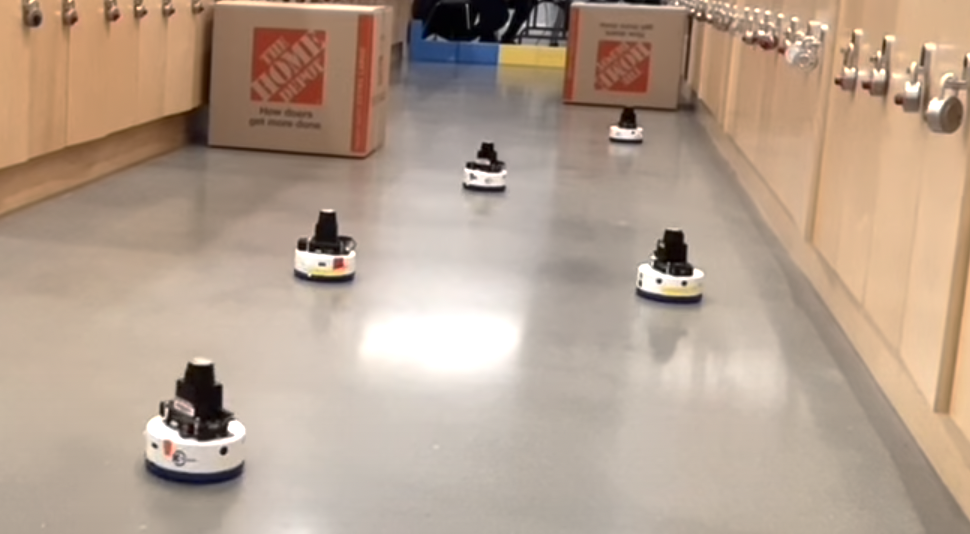}
    \includegraphics[width=5cm]{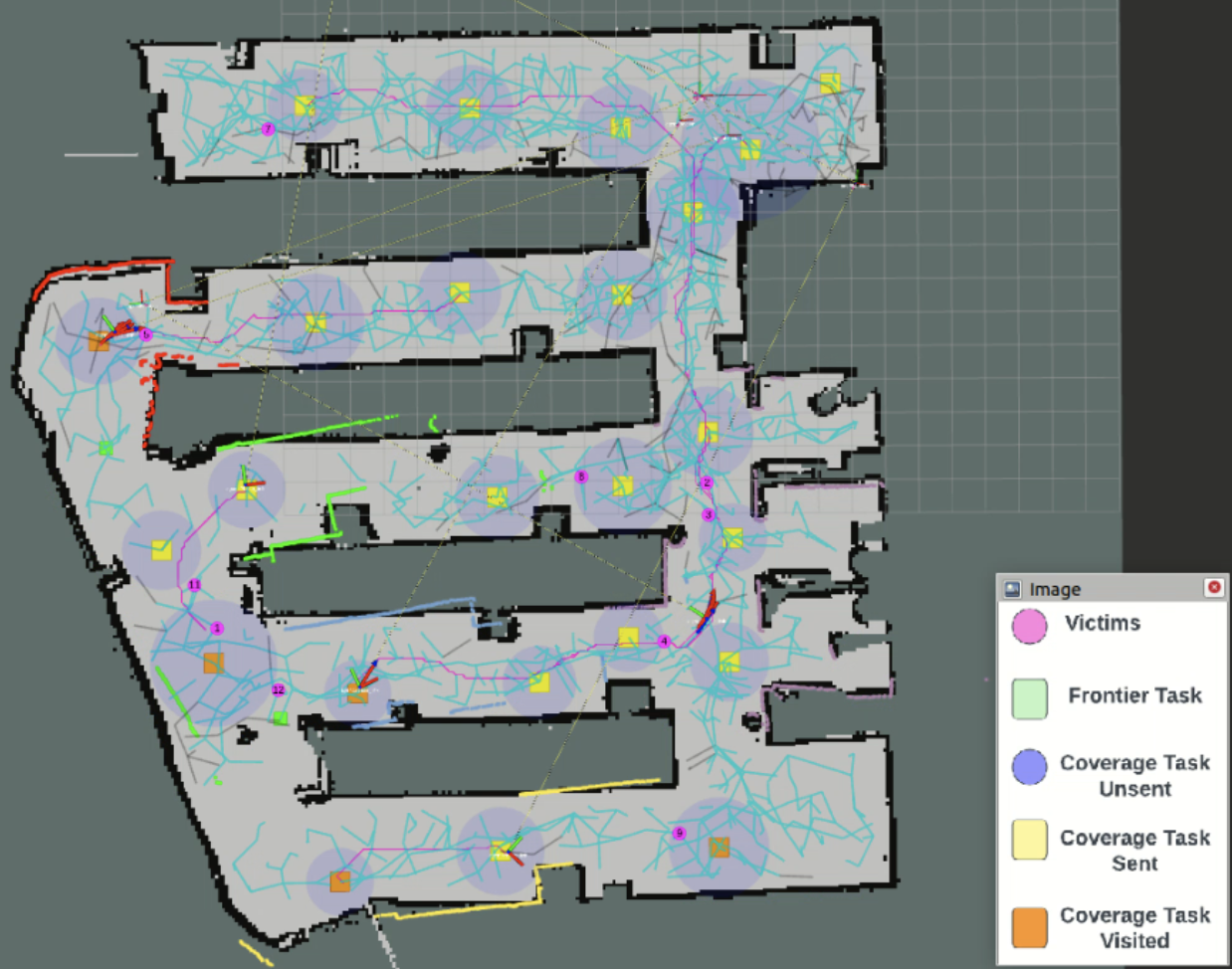}
    \caption{(Top) Real testing in an indoor environment (Bottom) System visualized in RViz during this real experiment}
    \label{fig:tepper_rviz}
\end{figure}
We conducted additional physical experiments with a fleet size of five robots to further evaluate the performance of our search and rescue system. The environment is a locker area with maze-like corridors perfect for testing multi-robot search depicted in Figure \ref{fig:tepper_rviz}. 

The robotic platform we are using is the Khepera IV robot. Each robot has a differential drive with wheel encoders (used for odometry), one URG Laser Range Finder, and one RGB camera.


Over five runs, HIGH had a 20.0\% better search efficiency $\epsilon$ and 30.1\% better SST than greedy NBV. These results align with our previous simulation results in Table \ref{tepper-sim-table}. The real-life experiments had slightly higher average $\epsilon$ and SST compared to simulation experiments. This sim-to-real gap is likely because of noisier sensor measurements, resulting in longer navigation times compared to simulation. 



\section{Conclusion}

This paper proposes novel algorithms to perform fast multi-robot search for victims in an unknown environment. The PGART planner introduces a unified approach for coverage-exploration planning for robotic search. It grounds the random trees on the pose graph to resolve the problem of excessive pruning of trees from loop closures and preventing backtracking into already visited areas. To account for the unknown distribution of victims in the environment, the HIGH allocator constructs a hierarchical information heuristic tree to dummy this distribution and prioritize search towards high information areas. Robots are allocated using a reward function that optimizes objectives of distance cost, information utility, information gain, and relative importance of exploration and coverage. Through simulation and real-life experiments, we demonstrated a 20\% more coverage efficiency compared to RRT exploration and greedy NBVP approach. We also make our algorithms and the supporting multi-robot autonomy stack open source. Future work includes extending this work to higher dimensions, including a longer planning horizon for HIGH, and implementing decentralized versions of our algorithms.

\section*{ACKNOWLEDGMENT}
This research was sponsored by the Advanced Agent - Robotics Technology lab at Carnegie Mellon University as a part of Masters in Robotics Systems Development program under the guidance of John Dolan and Dimi Apostolopoulos. We would like to thank Greg Armstrong, Akshaya Srinivasan, Joseph Campbell, Simon Stepputtis and the extended RoboSAR team for supporting this research in various ways.

\addtolength{\textheight}{0cm}   







\end{document}